\documentclass[10pt,twocolumn,letterpaper]{article}

\usepackage[accsupp]{axessibility}  
\usepackage{authblk}

\usepackage{cvpr}              

\usepackage[dvipsnames]{xcolor}

\definecolor{cvprblue}{rgb}{0.21,0.49,0.74}
\usepackage[pagebackref,breaklinks,colorlinks,citecolor=cvprblue]{hyperref}
\usepackage{multirow}
\usepackage{float}

\def\paperID{9549}
\def\confName{CVPR}
\def\confYear{2025}

\title{Single Domain Generalization for Few-Shot Counting \\via Universal Representation Matching}

\author{Xianing Chen \quad Si Huo \quad Borui Jiang \quad Hailin Hu\textsuperscript{*} \quad Xinghao Chen\thanks{Corresponding authors.} \\
\vspace{-0.6em}
\normalsize Huawei Noah’s Ark Lab. \\
\tt\small{sdxianing@gmail.com, \{huosi,jiangborui,hailin.hu,xinghao.chen\}@huawei.com}
}

\begin{document}
\maketitle

\begin{abstract}
	Few-shot counting estimates the number of target objects in an image using only a few annotated exemplars. However, domain shift severely hinders existing methods to generalize to unseen scenarios.  
	This falls into the realm of single domain generalization that remains unexplored in few-shot counting. 
	To solve this problem, we begin by analyzing the main limitations of current methods, which typically follow a standard pipeline that extract the object prototypes from exemplars and then match them with image feature to construct the correlation map. We argue that existing methods overlook the significance of learning highly generalized prototypes. 
	Building on this insight, we propose the first single domain generalization few-shot counting model, \underline{U}niversal \underline{R}epresentation \underline{M}atching, termed \textbf{URM}. Our primary contribution is the discovery that incorporating universal vision-language representations distilled from a large scale pretrained vision-language model into the correlation construction process substantially improves robustness to domain shifts without compromising in domain performance.
	As a result, URM achieves state-of-the-art performance on both in domain and the newly introduced domain generalization setting.
    \renewcommand{\thefootnote}{**}\footnotetext{Code is available at \url{https://github.com/jbr97/URM}.}
\end{abstract}
   
\section{Introduction}
\label{sec:intro} 

\begin{figure}[t]
	\centering
	\begin{subfigure}[b]{0.496\textwidth}
		\includegraphics[width=\linewidth]{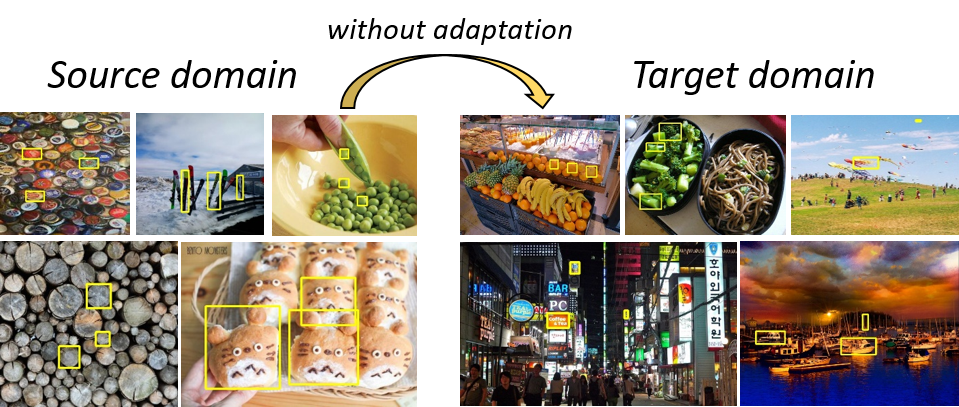}
		\caption{\textbf{domain generalization for few-shot counting}}
		\label{fig:dg}
	\end{subfigure}
	\begin{subfigure}[b]{0.457\textwidth}
		\includegraphics[width=\linewidth]{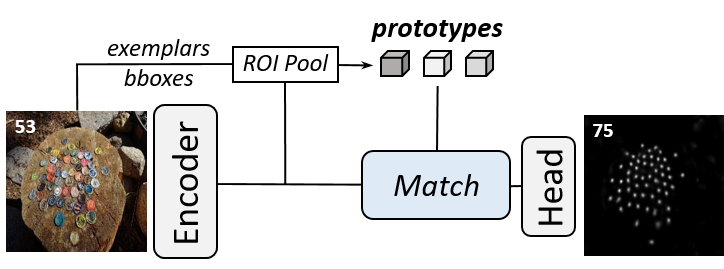}
		\caption{\textbf{extract-then-match}}
		\label{fig:suba}
	\end{subfigure}
	\begin{subfigure}[b]{0.496\textwidth}
		\includegraphics[width=\linewidth]{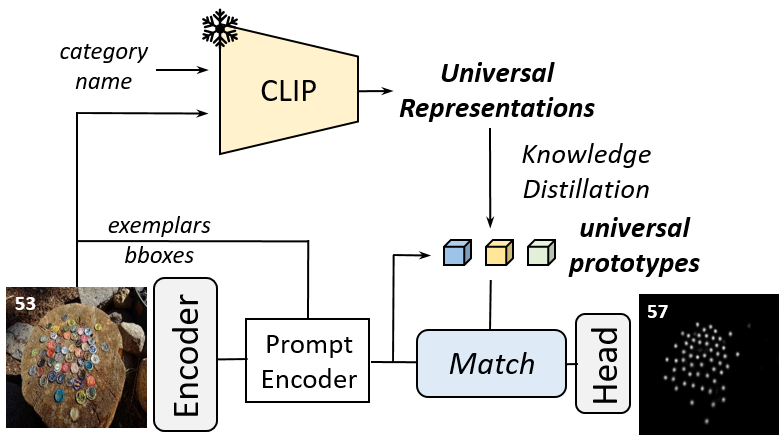}
		\caption{\textbf{universal representation matching}}
		\label{fig:subb}
	\end{subfigure}
    \vspace{-10pt}
	\caption{{Illustration of (a) domain generalization for few-shot counting, (b) the vanilla extract-then-match pipeline, and (c) our proposed universal representation matching.}}
	\label{fig_illu}
    \vspace{-15pt}
\end{figure}

Few-shot counting (FSC)~\cite{Ranjan2021LearningTC} is the task of estimating the number of target objects in an image using only a few annotated exemplars or without any exemplar at all.
Although many proposed approaches~\cite{You2022FewshotOC,Pelhan2024DAVEA} have achieved satisfactory results, they experience severe performance degradation when tested on data from unseen scenarios in real-world deployment due to the domain shift~\cite{Kouw2018AnIT} problem.
This shift arises from the deviation from the source domain training data to unseen target domain testing data, such as changes in camera position and weather condition \cite{Wang2019LearningFS,Zhu2022FineGrainedFD}, as well as biases introduced during data collection. For instance, images in the FSC147~\cite{Ranjan2021LearningTC} show each target object clearly while scenes in FSCD-LIVS~\cite{Nguyen2022FewshotOC} feature more occlusions and quality degradation as discussed in~\cite{Nguyen2022FewshotOC}. 

Although domain adaptation (DA)~\cite{Tzeng2017AdversarialDD} has been proposed to adapt the source domain distribution to that of the target domain by finetuning models with target domain data, the process is often laborious, and target domain data is not always available in practice, especially in the class-agnostic setting of FSC~\cite{Ranjan2021LearningTC}. 
In contrast, domain generalization (DG)~\cite{Zhou2021DomainGA} seeks to generalize the model to any unseen target domain without requiring target domain data and model updates as shown in Figure~\ref{fig:dg}, that is, to learn a generalized model by only source domain with a limited distribution. 
We consider in this work the unexplored and challenging field of single DG for few-shot counting to address the aforementioned challenges, where only one source domain---likely of narrow distribution---is used for training.

We start by analyzing why existing methods that follow the common \emph{extract-then-match} pipeline~\cite{Djukic2022ALO} illustrated in Figure~\ref{fig:suba} fail.
In this pipeline, exemplars prototypes are first extracted from the image feature. Next, candidates are generated by analyzing the correlation between prototypes and the image feature. Finally, the resulting correlation map is regressed into an object density map. 

The primary distinction among these methods lies in their approaches to constructing correlation, such as attention~\cite{Lin2021ObjectCY,Liu2022CounTRTG,Hobley2022LearningTC}, Siamese similarity~\cite{Djukic2022ALO,Ranjan2021LearningTC,Ranjan2022ExemplarFC}, and so on~\cite{Shi2022RepresentCA,You2022FewshotOC}. However, they all overlook the importance of prototypes in the construction process. \emph{Prototypes learned from the source domain with a narrow distribution inherently exhibit a similarly narrow distribution}. Intuitively, matching with such a limited representation in the unobserved target domain with a rather broad distribution inherently leads to sub-optimal performance. Therefore, enhancing the generalization of prototypes is essential for learning a domain-general few-shot counter.

The contrastive language-image pre-training model (CLIP)~\cite{Radford2021LearningTV} and its extensions~\cite{Yu2023TurningAC,Cheng2024TransferCF,Huang2024FROSTERFC,Zhou2021ExtractFD} have demonstrated remarkable generalization capacity in open-world down-stream tasks. The representation from CLIP can serve as \emph{universal representation} with domain-invariant information which demonstrate extraordinary performance across several distributions owing to the vast diversity of distributions seen during training~\cite{Ojha2023TowardsUF}.
\emph{Our finding indicates that matching with the universal vision-language representation distilled from the frozen CLIP model during the correlation construction process exhibit remarkable resilience to domain shift while without in domain performance degradation.} The operation is illustrated in Figure~\ref{fig:subb}.

Building on the discovery, we introduce the first domain-general few-shot counting model, \emph{\underline{U}niversal \underline{R}epresentation \underline{M}atching}, termed \textbf{URM}.
Specifically, we introduce universal vision and language prototypes to act as knowledge distiller~\cite{Touvron2020TrainingDI} to learn universal knowledge from the large-scale pre-trained CLIP model motivated by prompt knowledge distillation~\cite{Li2024PromptKDUP}, aiming to extend the narrow distribution of these prototypes. 
To adapt the V-L representations from CLIP for local objects matching, we minimally adjust the CLIP encoder while keeping its pre-trained weights frozen.

\begin{itemize}
	\item \emph{For the visual representation:} the class token focuses more on global image properties which may not suitable for capturing localized details~\cite{Dosovitskiy2020AnII}, and the image tokens are weak in local discriminability~\cite{Lan2024ClearCLIPDC}. Thus, we follow MaskCLIP~\cite{Zhou2021ExtractFD} to obtain class-specific segmentation, allowing us to extract locality object-level representations~\cite{Rasheed2022BridgingTG} which are then used to distill region-specific information into the visual prototypes.

	\item \emph{For the language representation:} we follow CuPL~\cite{Pratt2022WhatDA} to generate diverse customized prompts that leverage knowledge contained in LLMs~\cite{Achiam2023GPT4TR} as well as hand-written templates, to produce numerous descriptive sentences that capture important discriminative characteristics of the object categories. These prompts are subsequently encoded by the language tower to obtain universal language embeddings for distillation.
\end{itemize}

The learned universal V-L prototypes are then matched with the image feature through cross attention to construct the correlation map. As the distillation operation is conducted only during the training phase, URM remains as efficient as other methods that do not require external knowledge at inference time. 

To validate URM, we conduct extensive experiments on both cross and in domain setting on FSC147~\cite{Ranjan2021LearningTC} and FSCD-LVIS~\cite{Nguyen2022FewshotOC} datasets in both few-shot and zero-shot scenarios. The results demonstrate that URM performs excellently on both intra-dataset and cross domain settings. 

The main contributions of our work are as follows:

\begin{itemize}
	\item We introduce a new challenging setting of the single domain generalized few-shot counting task.
    And we point out that existing paradigm's inability to accurately count objects in new scenes stems from the learned narrow distribution prototypes.
	
	\item We introduce URM, the first narrow source domain-general few-shot counting model. Specifically, we propose to match universal V-L representations transferred from CLIP through knowledge distillation to construct the correlation map, a novel approach not explored in previous works.
	
	\item Our proposed URM outperforms state-of-the-art methods by a large margin in both the introduced narrow source domain general setting as well as in domain benchmark. 
\end{itemize}

\section{Related Works}
\label{sec:related}

\paragraph{Few-shot Counting.} Object counting refers to the task of estimating the number of specific objects present in an image, typically limited to a few predefined categories, \eg, human heads~\cite{Zhang2016SingleImageCC,Lian2021LocatingAC}, cells~\cite{Tyagi2023DeGPRDG}, cars~\cite{Wen2021DetectionTA}, and polyps~\cite{Zavrtanik2020ASA}. 
Few-shot counting addresses this limitation by given only a few annotated exemplars or without any exemplar at all. 
FamNet \cite{Ranjan2021LearningTC} proposed a pipeline that extracts appearance-based queries from
exemplars and matches them with image feature to infer the object counts. 
Then, BMNet+~\cite{Shi2022RepresentCA} enhanced localization by jointly learning representation with a non-linear similarity metric. 
CounTR~\cite{Liu2022CounTRTG} introduced cross attention-based interaction module to fuse image and exemplar features.
SAFE-Count~\cite{You2022FewshotOC} incorporated a feature enhancement module. 
LOCA~\cite{Djukic2022ALO} further refined this pipeline by iteratively adapting object prototypes based on exemplar appearance and shape.
DAVE~\cite{Pelhan2024DAVEA} proposed a post-process method for LOCA by generating a high-recall detection set and verifying the results to identify and remove outliers. 
All of these methods follow the same \emph{extract-then-match} paradigm and struggle with domain generalization.

\paragraph{CLIP-guided Counting.} Some studies also explored to count by leveraging CLIP. For instance,
\cite{Paiss2023TeachingCT} extended CLIP’s capabilities to counting but still failed when objects number is large. 
CrowdCLIP~\cite{Liang2023CrowdCLIPUC} mapped crowd patches to count text by constructing ranking-based text prompts to match the size-sorted crowd patches when category names were provided. 
Other works~\cite{jiang2023clip,amini2023open,Kang2023VLCounterTV,Xu2023ZeroShotOC,Wang2024LanguageGuidedZO,Zhu2024ZeroshotOC} explored enumerating objects given class names, creating a link between object categories and visual representations. Besides, some open-set methods~\cite{jiang2024t,ren2024grounding,yao2024detclipv3} achieve competitive results by leveraging CLIP as a bridge between V-L representations for interactive object detection. However, they require much more computational power and data for training, and their training classes cover the test classes.

\paragraph{Single Domain Generalization.} Since domain shift problem in FSC remains unexplored, we mainly focus on related approaches for crowd counting. 
Although some domain adaptation techniques exist~\cite{Zhu2023DAOTDA,Wu2021DynamicMA,Zhu2022FineGrainedFD}, they all inadequate for real-world application due to the lack of target data~\cite{Cai2023ExplicitIF,Gao2019DomainAdaptiveCC,Liu2020TowardsUC,Wang2019LearningFS,Zhu2022FineGrainedFD}. 
Domain generalization methods are primarily built on meta-learning algorithm MLDG~\cite{Li2017LearningTG}, which employs meta-train and meta-test set to simulate the domain shift between training and testing phases in reality to learn domain-invariant features. 
For instance, DCCUS~\cite{Du2022DomaingeneralCC} proposed a dynamic sub-domain division scheme, dividing the source domain into multiple sub-domains. 
Whitening-based methods~\cite{Pan2018TwoAO,Pan2019SwitchableWF,Choi2021RobustNetID} remove domain-specific style information by eliminating feature correlation to ensure each feature has unit variance.
However, these algorithms require a sufficiently broad source domain distribution.
MPCount~\cite{Peng2024SingleDG} proposed storing diverse density values for regression and reconstruct domain-invariant features using memory bank, eliminating the need for sub-domain partitioning. However, MPCount is still limited by the narrow source training data without accessing external knowledge.

\paragraph{Knowledge Distillation (KD).} \cite{Hinton2015DistillingTK} proposed to optimize a student model under the effective information transfer and supervision of a teacher model or ensemble. Although several methods have been proposed to learn knowledge from teacher network with high fidelity~\cite{Gou2020KnowledgeDA,Zhao2022DecoupledKD,Touvron2020TrainingDI}, we simply follow CLIP-KD~\cite{Yang2023CLIPKDAE} to use feature mimicry learning~\cite{Chen2022DearKDDE} which has been proved to be the most effective approach for CLIP knowledge distillation.

\begin{figure}[t]
	\centering
	\begin{subfigure}[b]{0.156\textwidth}
		\includegraphics[width=\linewidth]{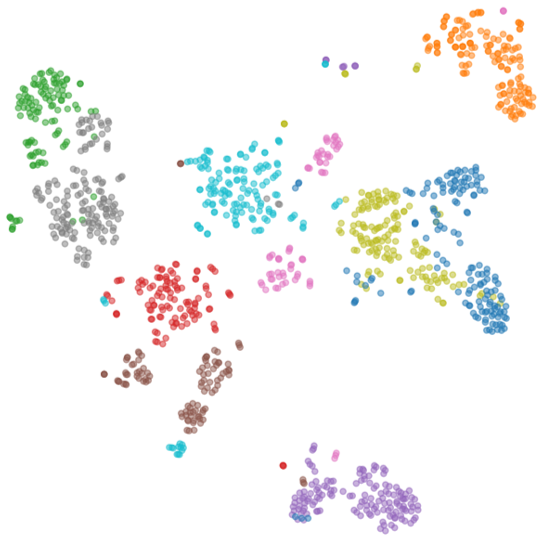}
		\caption{}
		\label{fig_tsnea}
	\end{subfigure}
	\begin{subfigure}[b]{0.156\textwidth}
		\includegraphics[width=\linewidth]{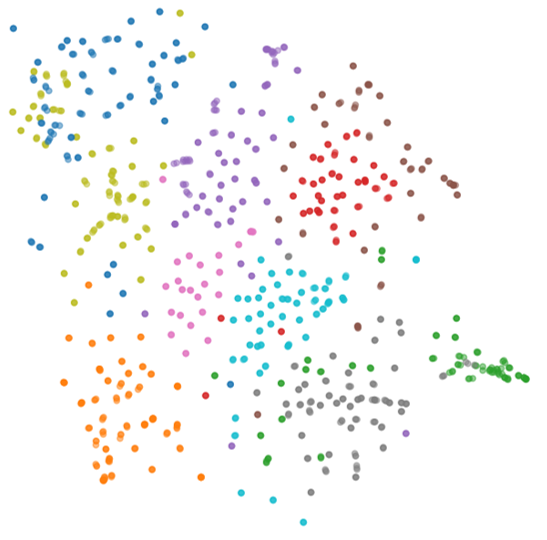}
		\caption{}
		\label{fig_tsneb}
	\end{subfigure}
	\begin{subfigure}[b]{0.156\textwidth}
		\includegraphics[width=\linewidth]{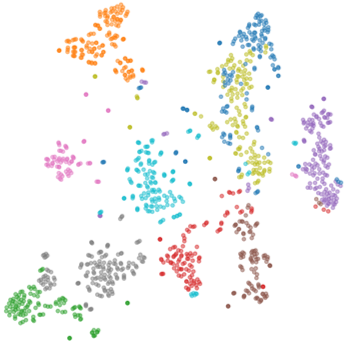}
		\caption{}
		\label{fig_tsnec}
	\end{subfigure}
	\caption{\textbf{t-SNE visualization of the prototypes feature space for different categories} from FSC147 by (a) the vanilla paradigm trained on FSC147, (b) the vanilla paradigm trained on FSCD-LVIS, and (c) our proposed URM trained on FSCD-LVIS.
  Note that the visualization is conduct on the test set where the object categories are disjoint from the train set. Best viewed in color.}
   \label{fig_tsne}
\end{figure}

\begin{figure*}[t]
	\centering
	\includegraphics[width=0.91\linewidth]{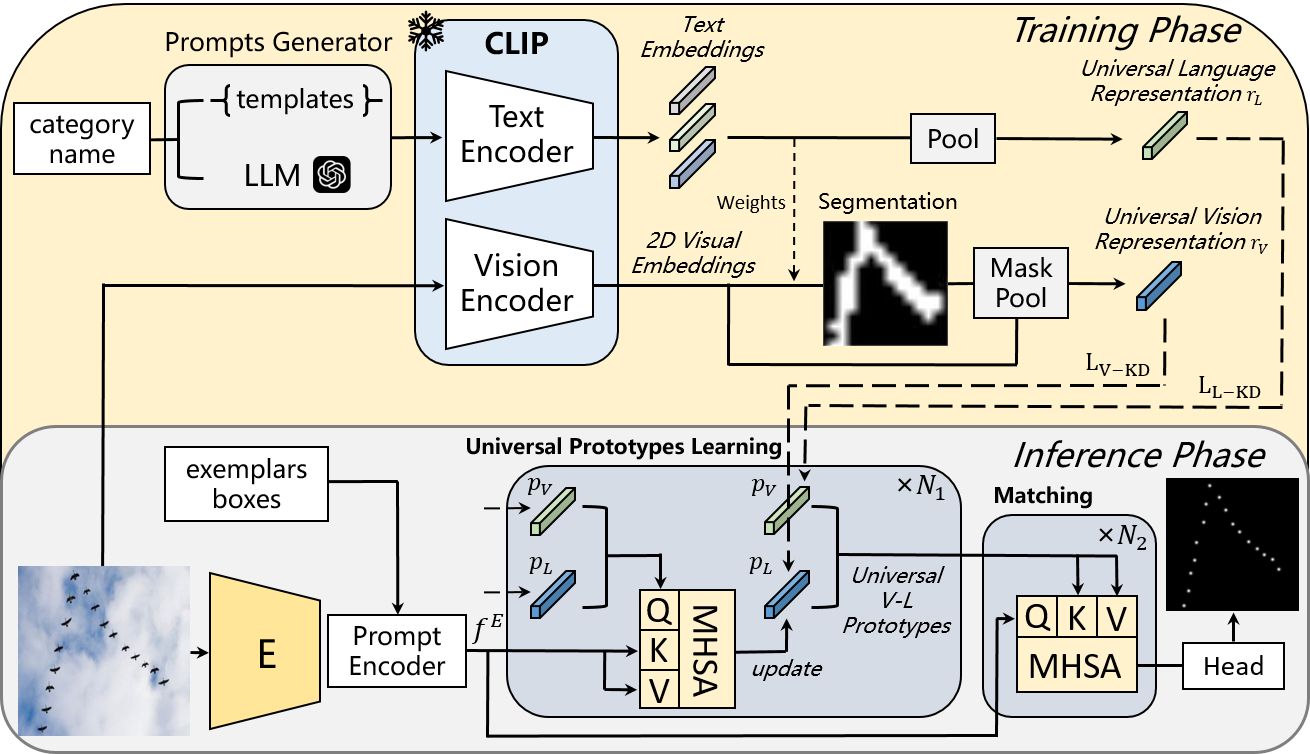}
	\caption{\textbf{The framework of our proposed URM.} The inference architecture is depicted in the gray part, where the learned prototypes are matched with the image feature through cross attention. The yellow part illustrates the universal V-L representations obtained from CLIP, which are distilled into the prototypes exclusively during the training phase.}
    \label{fig_method}
\end{figure*}

\section{Method}
\label{sec:method}
In this section, we first recap the preliminaries of CLIP, and then analyze why prior methods fail to generalize. Afterward, we introduce our proposed universal representation matching model.

\paragraph{Preliminary.} CLIP~\cite{Radford2021LearningTV}, trained on large-scale image-text pairs by image-level contrastive learning, has well demonstrated the potential for learning transferable knowledge and open-set visual concepts. 
CLIP consists of a pair of image encoder $\mathcal{V}(\cdot)$ and text encoder $\mathcal{T}(\cdot)$, both  of which are jointly trained to map input image and text into an unified representation space.
However, since CLIP is designed for image-level prediction, extracting local patch-level predictions from CLIP is non-trivial~\cite{Shi2023EdaDetOO}. 
MaskCLIP~\cite{Zhou2021ExtractFD} modifies
the image encoder by removing the last query and key embedding layers, as well as reformulating the value-embedding layer and the last linear layer into two respective convolutional layers, while keeping the text encoder unchanged. 
The text encoder then processes language prompts containing target classes, and the resulting text embedding for each class is used as a classifier to produce pixel-level mask predictions. 

\paragraph{Motivation.} Since the existing extract-then-match paradigm fails to generalize, we begin by analyzing the reason behind this undesirable phenomenon, which motivates our proposed solution.
After training on the source domain, we use t-SNE~\cite{Maaten2008VisualizingDU} to visualize the feature space of prototypes for different categories on the test set of FSC147 learned using both vanilla paradigm and URM. 
One noticeable observation is the clear boundary in Figure~\ref{fig_tsnea} using vanilla paradigm trained on FSC147, which corresponds to its well performance in the intra-dataset setting, even when the categories are disjoint from those in the training set.
In contrary, Figure~\ref{fig_tsneb} shows an unclear boundary on the target domain when the traditional paradigm is trained on dataset with large doamin gap, aligning with a significant drop in accuracy in the cross-domain setting.
The phenomenon suggests that simply learning prototypes within a narrow distribution is suboptimal for generalization.
The key, we believe, is that the matching process should occur in a feature space with well-defined open-set visual concept boundaries.
As shown in Figure~\ref{fig_tsnec}, prototypes from our URM---transferred from CLIP representation---better separate objects with unseen categories in unseen scenarios, supporting our hypothesis.

\subsection{Universal Representation Matching}
\paragraph{Overview.} The overall framework of our proposed URM is shown in Figure~\ref{fig_method}.
The input image $\mathbf{I} \in \mathbb{R}^{H \times W \times 3}$ is first encoded by the backbone network. 
Multi-scale features are extracted from the last three blocks, then resized to the same size of $h\times w$ and reduced to $d$ channels using a linear layer after concatenation.
Next, a set of $n$ selected exemplar bounding boxes are encoded into the image feature $\mathbf{f} \in \mathbb{R}^{h \times w \times d}$ via a prompt encoder.
After that, the proposed universal vision and language prototypes $p_v$ and $p_l \in \mathbb{R}^{n \times d}$ are initialized as learnable embeddings, and iteratively updated through interaction with the encoded image feature ${f^E} \in \mathbb{R}^{hw \times d}$ using $N_1$ cross attention layers~\cite{Vaswani2017AttentionIA}. During this process, the resulting prototypes are distilled from the universal V-L representations obtained from the frozen CLIP model, respectively.
Finally, the concatenated universal prototypes are correlated with  ${f^E}$ by $N_2$ cross attention layers to obtain a correlation map, which is then regressed into a density map $\mathbf{R} \in \mathbb{R}^{H \times W}$, whose number is summed to the count estimation, \ie, $c = sum( \mathbf{R} )$.

\begin{figure}[t]
	\centering
	\includegraphics[width=0.85\linewidth]{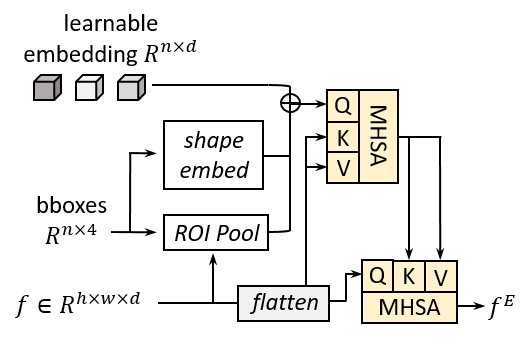}
	\caption{\textbf{Illustration of the prompt encoder.}}
    \label{fig_prompt_encoder}
\end{figure}

\paragraph{Prompt Encoder.} Given the extracted image feature $\mathbf{f}$, we use a two-way attention mechanism~\cite{Kirillov2023SegmentA} to encode exemplar prompts into $\mathbf{f}$, to indicate the object category to be counted. 
The exemplars are represented as a concatenated embedding composed of three elements: the height and width $R^{n \times 2}$ of bounding boxes $R^{n \times 4}$ which are mapped to shape embedding $R^{n \times d}$ by three linear layers with ReLU activation follow~\cite{Djukic2022ALO}, a learnable embedding group $R^{n \times d}$, and appearance tokens extracted from $\mathbf{f}$ using ROI Pooling~\cite{Ren2015FasterRT} . 
As illustrated in Figure~\ref{fig_prompt_encoder}, cross attention operations are then applied iteratively to update both the embedding and image feature, resulting in the category-aware image feature ${f^E}$. 

\paragraph{Universal Vision Representation Learning.} Since CLIP trained with image-level supervision lacks precise localization of objects and includes noise on non-relevant areas in its visual representation, 
we follow MaskCLIP~\cite{Zhou2021ExtractFD} to reformulate the value-embedding
layer of image encoder $\mathcal{V}(\cdot)$ to produce a dense attentive segmentation $\mathcal{M} \in \mathbb{R}^{h \times w}$.

Then, the universal vision representation $r_v$ containing fine-grained local image semantics is obtained by applying Mask Pooling~\cite{Chen2023OpenvocabularyPS} between the 2D visual embeddings of the image output from the last layer of CLIP vision encoder and the class-aware mask $\mathcal{M}$:

\begin{equation}
	r_v = \text{MaskPool}(\mathcal{V}(\mathbf{I}), \mathcal{M})
\end{equation}

Additionally, the vision prototype $p_v$ initialized as a learnable embedding $\mathbb{R}^{n \times d}$ is iteratively updated through $N_1$ cross attention layers with the encoded image feature, where $q=p_v$, $k=v={f^E}$.

Since distribution of the prototype learned from a narrow source domain is also narrow, we adopt feature mimicry loss following CLIP-KD~\cite{Yang2023CLIPKDAE} to transfer universal representation knowledge into the visual prototypes:

\begin{equation}
	\label{eq:1}
	\mathcal{L}_{V-KD}=\frac{1}{|\mathcal{B}|} \sum_{k=1}^{|\mathcal{B}|} \left\| {r_v}_k - {p_v}_k \right\|
\end{equation}
	
where $\mathcal{B}$ is the training batch, $r_v$ is the universal vision representation, and $p_v$ is the universal vision prototype. For simplicity, the linear projection layer for matching the dimension size of the CLIP embedding is omitted.

\paragraph{Universal Language Representation Learning.}  Hand-written templates like ``\textit{A photo of \{\}}'' completed with a category name are limited in describing details for fine-grained embedding~\cite{Pratt2022WhatDA}. 
Besides, writing high-performing prompt templates is labor-intensive and requires prior information about dataset contents. 
For example, ImageNet hand-written prompts~\cite{Radford2021LearningTV} like ``\textit{A toy \{\}}'' and ``\textit{A sculpture of \{\}}'' do not generalize well to other datasets like FSC-147.

We thus follow CuPL~\cite{Pratt2022WhatDA} to further extend templates with customized prompts generated by LLM that contain important discriminating characteristics which can best represent an object category in natural language. The templates and prompts for LLM, \ie, GPT-4~\cite{Achiam2023GPT4TR} in this work, are detailed in Appendix. 

The generated prompts for each category are then encoded by CLIP’s text encoder $\mathcal{T}(\cdot)$. The corresponding text embeddings are averaged to produce the resulting universal text representation $r_t$. 

Similar to vision prototype learning, the language prototype $p_l$ is iteratively updated through interactions with $f^E$ using the same cross attention layers.
And the language distillation loss is given by:

\begin{equation}
	\label{eq:2}
	 \mathcal{L}_{L-KD}=\frac{1}{\mathcal{B}} \sum_{k=1}^{\mathcal{B}} \left\| {r_l}_k - {p_l}_k \right\|
\end{equation}

where $r_l$ is the universal language representation, $p_l$ is the universal language prototype. The linear projection layer is also applied to the language prototype. 

\paragraph{Universal Representation Matching.} After obtaining the universal V-L prototypes for the object category in the image, we can construct a correlation between the concatenated prototypes and image feature. 
Although various matching approaches have been proposed as discussed, we focus on the essential matching elements within the correlation construction process. 
Specifically, we apply $N_2$ cross attention layers with $\text{q}={f^E}, \text{k}=\text{v}=\text{concat}(p_v, p_l)$ to obtain the correlation map $\mathbb{R}^{hw \times d}$.

Finally, a regression head predict the final 2D density map $\mathbf{R} \in \mathbb{R}^{H \times W}$ from the correlation map.

\paragraph{Loss Function.} The model is trained with $\ell_2$ loss between the predicted density map $\mathbf{R}$ and the ground-truth map $\mathbf{G}$, normalized by the number of objects:

\begin{equation}
	\mathcal{L}_{density}= \frac{1}{2\mathcal{B}} \sum_{k=1}^{\mathcal{B}} \frac{1}{N_k} \|\boldsymbol{G}_k-\boldsymbol{R}_k\|_2^2 
\end{equation}

where $N_k$ represents the number of objects in image $k$. The overall training loss is thus a combination of the density and the distillation losses from Equation~(\ref{eq:1})~(\ref{eq:2}):

\begin{equation}
	\label{eq:loss} 
	\mathcal{L}= \mathcal{L}_{density}+ \alpha \mathcal{L}_{V-KD} + (1 - \alpha) \mathcal{L}_{L-KD}
\end{equation}
	
where $\alpha$ is the distillation weight parameter to balance the contributions of the vision and language representations in the universal prototypes.

\begin{table*}[t] 
	\centering
	\begin{tabular}{c|cc|cc|cc|cc}
		\hline
		\multicolumn{1}{c|}{ Setting } & \multicolumn{4}{c|}{ Few-shot } & \multicolumn{4}{c}{ Zero-shot } \\
		\hline 
		\multicolumn{1}{c|}{ Source $\rightarrow$ Target } & \multicolumn{2}{c|}{ A $\rightarrow$ B } & \multicolumn{2}{c|}{B $ \rightarrow $ A} & \multicolumn{2}{c|}{A $\rightarrow$ B } & \multicolumn{2}{c}{ B $\rightarrow$ A } \\
		\hline 
		Metric & MAE & RMSE & MAE & RMSE & MAE & RMSE & MAE & RMSE \\
		\hline 
		\multicolumn{9}{c}{\textit{Vanilla Methods}} \\
		\hline
		FamNet (CVPR21)~\cite{Ranjan2021LearningTC} & 29.98 & 48.95 & 42.74 & 118.47 & / & / & / & / \\
		BMNet+ (CVPR22)~\cite{Shi2022RepresentCA} & 37.86 & 68.94 & 36.53 & 116.02  & / & / & / & / \\
		CounTR (BMVC22)~\cite{Liu2022CounTRTG} & 34.76 & 54.39 & 33.24 & 107.34 & / & / & / & / \\
		LOCA (ICCV23)~\cite{Djukic2022ALO} & 30.63 & 44.33 & 29.13 & 88.06 & 31.17 & 46.98 & 34.13 & 113.72 \\
		DAVE (CVPR24)~\cite{Pelhan2024DAVEA} & 29.67 & 45.28 & 29.51 & 86.13 & 30.42 & 47.03 & 35.40 & 115.81 \\
		\hline
		\multicolumn{9}{c}{\textit{Domain Generalization Methods}} \\
		\hline
		ISW (CVPR21)~\cite{Choi2021RobustNetID} & 35.95 & 51.83 & 36.41 & 124.09 & / & / & / & / \\
		GS (ICCV21)~\cite{Mansilla2021DomainGV} & 30.55 & 47.57 & 29.83 & 98.42 & / & / & / & / \\
		DCCUS (AAAI23)~\cite{Du2022DomaingeneralCC} & 29.02 & 43.55 & 28.46 & 83.32 & 30.39 & 44.87 & 34.29 & 109.20 \\
		MPCount (CVPR24)~\cite{Peng2024SingleDG} & \underline{25.11} & \underline{41.32} & \underline{22.07} & \underline{80.17} & \underline{27.68} & \underline{44.58} & \underline{30.99} & \underline{102.18} \\
		\hline
		URM (Ours) & $\textbf{21.87}$ & $\textbf{38.42}$ & $\textbf{21.17}$ & $\textbf{73.42}$  & $\textbf{23.54}$ & $\textbf{39.93}$ & $\textbf{27.49}$ & $\textbf{94.57}$ \\
		\hline
	\end{tabular}
	\caption{\textbf{Comparison with the state-of-the-art methods} on FSC-147 (A) and FSCD-LVIS (B).}
	\label{tab_result}
\end{table*}

\section{Experiments}
\label{sec:exp}
In this section, we evaluate the effectiveness of URM, demonstrating how universal representation matching enhances the generalization and performance of few-shot counting models.  

\subsection{Experimental setting}
\label{sec:setting}
\paragraph{Implementation details.} 
We first resize input images to $H=W=512$ and apply standard data augmentations such as tiling, horizontal flipping and color jitter. 
The model uses an Imagenet pre-trained Resnet50 backbone, with features from the final three blocks upsampled to $h=w=64$. 
The backbone network parameters are frozen follow~\cite{Ranjan2021LearningTC,Djukic2022ALO}, while all other parameters are trained for 150 epochs. 
The resulting features are projected into $d=256$ channels via an linear layer.
The distillation weight $\alpha$ in Equation~(\ref{eq:loss}) is set to 0.9.
The number of cross attention layers is $N_1=N_2=3$.
The regression head consists of
three 3 × 3 convolutional layers with 128, 64 and 32 feature channels, each followed by a Leaky ReLU, a 2× bilinear upsampling, and a linear layer, consistent with the LOCA configuration.
We use AdamW optimizer with the fixed learning rate $10^{-4}$ and a weight decay of $10^{-2}$. Gradient clipping with maximum norm of $0.1$ is applied. URM is trained on a single NVIDIA V100 GPU with batch size of 4.

\paragraph{Datasets.} We evaluate our method on two mainstream datasets: 
FSC-147~\cite{Ranjan2021LearningTC} and FSCD-LVIS~\cite{Nguyen2022FewshotOC}. 
Specifically, FSC-147 contains 6135 images of 147 object categories, split into training, validation and test sets consisting of 3659, 1286 and 1190 images respectively. Images in FSC-147 are collected from the Internet. The sets of object categories present in each split are disjoint. 
FSCD-LVIS contains 6195 images spanning 372 classes, which are extracted from the LVIS dataset~\cite{Gupta2019LVISAD}. 
Each image annotation consists of three bounding boxes for exemplar objects within each category.

Let $\mathcal{S} \rightarrow \mathcal{T}$ denote the case where $\mathcal{S}$ is the source domain and $\mathcal{T}$ is the target domain. The datasets are utilized in experiments under two cross-domain settings:  the full training dataset of one is regarded as the source domain, while other entire test dataset is regarded as the target domain, \ie, FSC-147 $\rightarrow$ FSCD-LVIS and FSCD-LVIS $\rightarrow$ FSC-147.

\paragraph{Evaluation metrics.} We evaluate our method using two commonly employed metrics in object counting: mean absolute error (MAE) and root mean square error (RMSE), between estimated and labeled counts for each image: 

\begin{equation}
M A E=\frac{1}{N} \sum_{i=1}^N\left|c_i-\hat{c_i}\right|, R M S E=\sqrt{\frac{1}{N} \sum_{i=1}^N\left(c_i-\hat{c}_i\right)^2}
\end{equation}

where $N$ is the number of test images, $c_i$ is the ground truth count for the $i$-th image, and $\hat{c_i}$ is the predicted count. Lower values of both metrics indicate better performance.

\subsection{Comparison with the state of the art}
\label{sec:result}
In this section, we compare our URM with state-of-the-art methods on different benchmarks.   
The selected baselines can be grouped into two categories:

\begin{itemize}
	\item \emph{Vanilla Methods}: FamNet~\cite{Ranjan2021LearningTC}, BMNet+~\cite{Shi2022RepresentCA}, CounTR~\cite{Liu2022CounTRTG}, LOCA~\cite{Djukic2022ALO}, and DAVE~\cite{Pelhan2024DAVEA}.

	\item \emph{Domain Generalization Methods}: ISW~\cite{Choi2021RobustNetID,Peng2024SingleDG}, GS~\cite{Mansilla2021DomainGV}, DCCUS~\cite{Du2022DomaingeneralCC}, and MPCount~\cite{Peng2024SingleDG}. 
\end{itemize}

All these DG methods are originally designed or adapted for crowd counting, we add the prompt encoder after feature extraction based on their implementation. Besides, GS is a combination of Gradient Surgery~\cite{Mansilla2021DomainGV} and LOCA~\cite{Djukic2022ALO}. 

Table~\ref{tab_result} presents the results of domain generalized few-shot counting. 
Our URM surpasses all the methods significantly, providing compelling evidence for the effectiveness of our designation. 
In contrast, sub-domain division methods like DCCUS and GS fail to achieve satisfactory results, as they struggle to produce meaningful sub-domains when the source domain distribution is narrow and the domain gap is large. 
Additionally, the whitening-based method ISW is even harmful because useful information is eliminated through the feature statistics normalization operation. 
Although MPCount introduced a single memory bank for regression does not require sub-domain partitioning, its generalization ability remains far inferior to our method.

We further evaluate our method on the zero-shot setting with no user annotation, the results are shown in the right part of Table~\ref{tab_result}. 
We follow~\cite{Djukic2022ALO} to take a minor modification of the prompt encoder by using only the learnable embedding to represent the exemplar prompt. 
The results confirm that our proposed URM effectively enhances the generalization ability of prototypes, enabling accurate count estimation even in the challenging case without manually annotated exemplar.

\begin{table}[t] 
	\centering
	\setlength{\tabcolsep}{12pt}
	\begin{tabular}{c|cc}
		\hline 
		Method & MAE & RMSE \\
		\hline 
		Baseline & 30.15 & 48.04 \\
		\hline 
		\multicolumn{3}{c}{\textit{Language Distillation}} \\
		\hline 
		Naive Template & 25.44 & 42.59 \\
		Prompt Generator & 23.83 & 41.03 \\
		\hline 
		\multicolumn{3}{c}{\textit{Vision Distillation}} \\
		\hline 
		\texttt{[CLS]} Token & 23.53 & 41.12 \\
		Global Pooling & 22.94 & 39.44 \\
		Mask Pooling & 21.87 & 38.42 \\
		\hline
	\end{tabular}
	\caption{\textbf{Ablation of the designed elements.} 
	}
	\label{tab_abla}
\end{table}

\subsection{Analysis and Ablations}
\label{sec:abla}

\paragraph{Ablation of the designed elements.} We first ablate the important elements of our designation under the FSC-147 (A) $\rightarrow$ FSCD-LVIS (B) few-shot setting. 
The baseline model shares the same architecture as URM in the inference phase but without knowledge distillation. The result shown in the first row of Table~\ref{tab_abla} demonstrates that simply altering architecture, \ie, matching with learned prototypes that lack knowledge transfer and thus have a narrow distribution, introduces no benefit for generalization. 

In contrast, the method can achieve comparable performance to the SOTA method MPCount by using only language knowledge from CLIP with a naive template. 
Since the naive template ``\textit{A photo of \{\}}'' is too limited to fully represent a category, adding more templates and customized language prompts by the prompt generator further improves the model performance. 

Subsequently, incorporating visual representation via global average pooling from the 2D vision tokens output from the last layer improve performance, demonstrating that the vision representation is complementary to language knowledge.
Using the \texttt{[CLS]} token which captures global description results in lower performance.
Finally, by introducing more precise locality representation as well as noise removal through mask pooling, our model reaches a MAE of $21.87$ as shown in the last row of Table~\ref{tab_abla}, outperforming the baseline method with $27.5\%$.

\paragraph{Ablation of distillation weight.} The distillation weight $\alpha$ in Equation~(\ref{eq:loss}) controls the balance between visual and language knowledge contributions. In this part, we conduct an ablation study on the weight to fully understand its impact. As shown in Table~\ref{tab_weight}, when $\alpha=0$ or $\alpha=1$, where only vision or language knowledge is introduced respectively, the performance is inferior to other settings, demonstrating that \textit{V-L representations complement each other}. Setting $\alpha=0.9$ yields the best performance, indicating that \textit{language representation from CLIP, which has been aligned to visual space, plays an influential role}.

\begin{table}[t] 
	\centering
	\setlength{\tabcolsep}{5pt}
	\begin{tabular}{c|ccccccc}
		\hline 
		$\alpha$ & 0 & 0.25 & 0.5 & 0.75 & 0.9 & 1\\
		\hline 
		MAE & 26.54 & 23.44 & 23.30 & \underline{22.37} & \textbf{21.87} & 23.83 \\
		RMSE & 42.82 & 39.87 & 39.48 & \underline{38.92} & \textbf{38.42} & 41.03 \\
		\hline
	\end{tabular}
	\caption{\textbf{Ablation of the distillation weight.}}
	\label{tab_weight}
\end{table}

\begin{table}[t] 
	\centering
    \setlength{\tabcolsep}{5pt}
	\begin{tabular}{c|cccccc}
		\hline 
		$N_1$ & 2 & 3 & 3 & 3 & 4\\
		\hline 
		$N_2$ & 3 & 2 & 3 & 4 & 3\\
		\hline 
		MAE & 22.77 & 22.23 & \textbf{21.87} & 23.58 & 23.45 \\
		RMSE & 39.69 & 39.12 & \textbf{38.42} & 41.22 & 39.38 \\
		\hline
	\end{tabular}
	\caption{\textbf{Ablation of the number of layers}, where $N_1$ is the universal prototypes learning layers, and $N_2$ is the universal representation matching layers.}
	\label{tab_layers}
\end{table}

\paragraph{Ablation of the number layers.} We then ablate the hyper-parameters related to the number of layers. The universal prototypes learning layer number $N_1$ determines how many layers are required to absorb the universal knowledge transferred from CLIP. And the universal representation matching layer number $N_2$ defines how many layers are needed to learn an effective correlation map during matching in the hidden space. The results in Table~\ref{tab_layers} show that $N_1=N_2=3$ achieves the best performance.

\begin{table}[t] 
	\centering
	\setlength{\tabcolsep}{4pt}
	\begin{tabular}{c|c|c|cc}
		\hline 
		Model & Architecture & ACC & MAE & RMSE \\
		\hline 
		OpenAI CLIP~\cite{Radford2021LearningTV} & ViT-B/16 & 68.3 & \underline{21.87} & 38.42 \\
		OpenAI CLIP~\cite{Radford2021LearningTV} & ViT-L/14 & 75.5 & 22.63 & 38.95 \\
		OpenAI CLIP~\cite{Radford2021LearningTV} & RN50$\times$16 & 70.5 & 23.15 & 39.63 \\
		OpenCLIP~\cite{Cherti2022ReproducibleSL} & ConvNeXt-B & 71.5 & 22.78 & 39.10 \\
		SigLIP~\cite{Zhai2023SigmoidLF} & ViT-B/16 & 73.4 & \textbf{21.33} & \textbf{37.85} \\
		EVA-02-CLIP~\cite{Sun2023EVACLIPIT} & ViT-B/16 & 74.7 & 21.86 & \underline{38.12} \\
		\hline
	\end{tabular}
	\caption{\textbf{Ablation of different universal V-L representation sources}, where ACC represents the ImageNet zero-shot classification performance of the CLIP model.}
	\label{tab_source}
\end{table}

\paragraph{Ablation of different representation sources.} In this part, we investigate the effect of different representation learning sources in Table~\ref{tab_source}. 
The first row presents the results of our default setting which uses the OpenAI ViT-B/16 CLIP model.
Increasing the model size leads to a performance drop due to the large dimension gap between the CLIP model and URM as seen in the second row.
Besides, models with convolutional backbones, such as ResNet50 and ConvNeXt, perform poorly in comparison despite achieving higher ImageNet accuracy, suggesting that ViT architecture is better suited for representation learning in our task.
Finally, the results in the last two lines indicate that using a stronger CLIP teacher model can further enhance our model's performance. However, this improvement also does not directly correlate with ImageNet accuracy, aligning with the observations in~\cite{Laurenon2024WhatMW}.

\subsection{Performance on in domain benchmark}
\label{sec:inter_result}
In this section, URM is compared with the state-of-the-art
methods on in domain benchmark. The three-shot evaluation results on FSC147 dataset are presented in Table~\ref{tab_tradition}. 
While demonstrating strong generalization, URM outperforms state-of-the-art few-shot counting models in terms of in-domain performance.
Additionally, DAVE~\cite{Pelhan2024DAVEA} proposed a detect-and-verify paradigm for the existing regression-based methods like LOCA to mitigate false activations on objects of other categories. 
Since DAVE is a general method for merging the benefits of both density and detection-based approaches, we also extend our method by incorporating this paradigm. Although inductive biases introduced by DAVE are insufficient to handle complex distribution gap in domain generalization setting (as show in Table~\ref{tab_result}), the resulting method still surpasses other methods on the traditional in domain benchmark, as demonstrated in the last row of Table~\ref{tab_tradition}. 

\subsection{Generating language prompts with LVLMs}
In this section, we explore the use of large vision-language models to generate language prompts, enabling our method to train without predefined category names.
However, existing proprietary and public LVLMs struggle with image classification especially in open-world setting where the class list is unknown~\cite{Zhang2024WhyAV}, making them difficult to accurately fill templates in the prompt generator.
Fortunately, customized descriptive prompts with details are effective~\cite{Pratt2022WhatDA} as shown in Section~\ref{sec:abla} and Appendix.
As a result, we choose Claude3-Opus~\cite{Claude} and GPT4~\cite{Achiam2023GPT4TR} to generate language prompts which have demonstrated the best open-world classification performance among existing VLMs~\cite{Zhang2024WhyAV}. 
We replace ``\textit{\{\}}'' in our template prompts with ``\textit{The category of the most numerous objects in the image}'' to prompt the VLMs. 
For example, ``\textit{Describe what a(n) \{\} looks like?}'' becomes ``\textit{Describe what the category of the most numerous objects in the image looks like?}''.
The results in Table~\ref{tab_vlm} show that our method can successfully generalize to scenarios where category names are not provided during training.

\begin{table}[t] 
	\centering
	\setlength{\tabcolsep}{2.8pt}
	\begin{tabular}{c|cc|cc}
		\hline 
		Split & \multicolumn{2}{c|}{Validation set} & \multicolumn{2}{c}{Test set} \\
		\hline 
		Method & MAE & RMSE & MAE & RMSE  \\
		\hline 
		FamNet (CVPR21)~\cite{Ranjan2021LearningTC} & 23.75 & 69.07 & 22.08 & 99.54 \\
		BMNet+ (CVPR22)~\cite{Shi2022RepresentCA} & 15.74 & 58.53 & 14.62 & 91.83 \\
		VCN (CVPR22)~\cite{Ranjan2022VicinalCN} & 19.38 & 60.15 & 18.17 & 95.60 \\
		CountTR (BMVC22)~\cite{Liu2022CounTRTG} &  13.13 & 49.83 & 11.95 & 91.23 \\
		SAFECount (WACV23)~\cite{You2022FewshotOC} & 15.28 & 47.20 & 14.32 & 85.54 \\
		PseCo (CVPR24)~\cite{Huang2023PointSA} & 15.31 & 68.34 & 13.05 & 112.86 \\
		\hline
		LOCA (ICCV23)~\cite{Djukic2022ALO} & 10.24 & 32.56 & 10.79 & 56.97 \\
		DAVE (CVPR24)~\cite{Pelhan2024DAVEA} & \underline{8.91} & \underline{28.08} & \underline{8.66} & \underline{32.36} \\
		\hline
		URM (Ours) & 10.63 & 32.01 & 10.15 & 54.83 \\
		URM+DAVE~\cite{Pelhan2024DAVEA} & $\mathbf{8.47}$ & $\mathbf{26.51}$ & $\mathbf{8.32}$ & $\mathbf{32.05}$ \\
		\hline
	\end{tabular}
	\caption{\textbf{Performance comparison on in domain benchmark.}}
	\label{tab_tradition}
\end{table}

\begin{table}[t] 
	\centering
	\begin{tabular}{c|cc}
		\hline 
		LVLM & MAE & RMSE \\
		\hline 
		Claude3~\cite{Claude} & 22.58 & 38.96 \\
		GPT4~\cite{Achiam2023GPT4TR} & 22.45 & 38.75 \\
		\hline
	\end{tabular}
	\caption{\textbf{Results of generating language prompts using LVLMs when category names are not given.}}
	\label{tab_vlm}
\end{table}

\section{Conclusion}
This paper proposes URM to address single domain generalization for few-shot counting with narrow source distribution. 
URM is motivated by the unique challenge that existing methods struggle to learn high generalization prototypes. 
To address this issue, we propose a universal representation matching paradigm that incorporating universal vision-language prototypes distilled from the universal representations from CLIP model, which has seen vast diversity of distributions during training.
Matching with these universal prototypes leads to high generalization and superior performance.
Extensive experiments on both cross and in domain settings show that URM significantly outperform the state-of-the-art.

\newpage
{
    \small
    \bibliographystyle{ieeenat_fullname}
    \bibliography{main}
}

\clearpage
\setcounter{page}{1}
\maketitlesupplementary

\section{More Analysis}
\paragraph{Comparison with prototypes generated by CLIP.} 
URM learns universal knowledge through distillation only during the training phase,  making it as efficient as other methods during inference. 
In this section, we further compare our method with generating prototypes by CLIP vision encoder directly, which introduces more parameters. 
Specifically, the input image is encoded by the CLIP vison tower, and prototypes are obtained via global average pooling from the 2D visual tokens. 
Since obtaining category names during testing is unrealistic, we rely solely on the vision prototype. 
For fair comparison, we compare this approach with URM using only vision prototype, termed URM-V.
The results in Table~\ref{tab_clip_enc_1} demonstrate that our method can learn useful representation as well as achieve better performance.

\begin{table}[t] 
	\centering
    \setlength{\tabcolsep}{6pt}
	\begin{tabular}{c|ccc}
		\hline 
		Method & MAE & RMSE \\
		\hline 
		CLIP Enc. & 27.32 & 43.28 \\
        URM-V & 26.54 & 42.82 \\
		\hline
	\end{tabular}
	\caption{\textbf{Comparison with generating visual prototypes by using CLIP vision encoder directly.}}
	\label{tab_clip_enc_1}
\end{table}

\paragraph{CLIP as Backbone \textit{E}.} The result of URM w.o. distillation with the backbone changed to CLIP in Table~\ref{tab_clip_enc} shows that this setting is inferior. This is mainly because CLIP with global supervision lacks fine-grained spatial details that are crucial for object counting. While the universal representation in CLIP does provide some improvements, it is not sufficient to compensate for the loss of spatial information. And the improvements primarily stem from the language representation.

\begin{table}[]
  \centering
  \setlength{\tabcolsep}{1pt}
  \begin{tabular}{c|cc|cc}
    \toprule
    Metric&MAE &MSE &MAE &MSE \\
    \midrule
    Task & \multicolumn{2}{c|}{A $\rightarrow$ B} & \multicolumn{2}{c}{A $\rightarrow$ A} \\
    \midrule
    CLIP(RN50$\times$16) & 28.8(-1.4) & 45.7(-2.3) & 11.7(+0.9) & 61.4(+4.2) \\
  \bottomrule
  \end{tabular}
  \caption{\textbf{Result of URM using CLIP as backbone and without universal knowledge distillation.}}
  \label{tab_clip_enc}
\end{table}

\section{Results of Domain Adaptation Methods}
In this section, we  further conduct experiments on domain adaptation methods. 
The results are shown is Table~\ref{tab_adapt} and are provided only for reference as a part of data from target domain is visible during adaptation.
We select SE CycleGAN~\cite{Wang2019LearningFS} and DAOT~\cite{Zhu2023DAOTDA}, both of which are domain adaptation methods for crowd counting and have released code. 
Specifically, SE CycleGAN learns to translate scenes from one domain to another, aiming to reduce low-level differences.
DAOT aligns domain-agnostic factors between domains via an aligned optimal transport strategy.
It is worth noting that even part of the target data is provided, these DA methods cannot achieve results as good as those in other fields~\cite{Peng2024SingleDG,Liu2023StructuralRI}, when data categories are disjoint in the context of class-agnostic setting of FSC, highlighting the importance the domain generalized FSC. Additionally, the poor performance of SE CycleGAN suggests that low-level image style is not the influencing factor contributing to domain shift.

\begin{table}[t] 
	\centering
    \setlength{\tabcolsep}{5.5pt}
	\begin{tabular}{c|cc|cc}
        \toprule
	\multicolumn{1}{c|}{ Source $\rightarrow$ Target } & \multicolumn{2}{c|}{ A $\rightarrow$ B } & \multicolumn{2}{c}{B $ \rightarrow $ A} \\
		\hline 
		Metric & MAE & RMSE & MAE & RMSE \\
        \hline
		SE CycleGAN~\cite{Wang2019LearningFS} & 32.48 & 50.77 & 34.63 & 115.97 \\
		DAOT~\cite{Zhu2023DAOTDA} & 22.79 & 39.33 & 20.82 & 73.44 \\
        \bottomrule
	\end{tabular}
	\caption{\textbf{Results of the domain adaptation methods.}}
	\label{tab_adapt}
\end{table}

\section{Details of the Prompt Generator}
\label{supp:generator}
In this section, we first detail the hand-written templates and prompts for LLMs we used in Table~\ref{tab_cupl}. Then, we perform an ablation study on the prompts for language representation in Table~\ref{tab_temp}. Although using the naive template only achieves satisfactory performance, we find that applying all the Imagenet prompts used in CLIP leads to poor performance, due to the dataset distribution gap discussed above. 
Thus, we only use the general templates which yield better performance. 
Additionally, it is worth noting that we append "within 20 words" to some prompts for LLMs, as GPT tends to refuse to answer questions that describe low-detail objects.
Finally, URM-L with language representation learning alone achieves the best result with our prompt generator.

\begin{table*}[htbp] 
	\centering
	\begin{tabular}{c|c}
		\multirow{5}{*}{Templates} &  A photo of a \{\texttt{category name}\}. \\
		& A photo of \{\texttt{number}\} \{\texttt{category name}\}. \\
		& A bad photo of a \{\texttt{category name}\}. \\
		& A photo of many \{\texttt{category name}\}. \\
		& A low resolution photo of the \{\texttt{category name}\}. \\
        & A photo of a hard to see \{\texttt{category name}\}. \\
        & A cropped photo of a \{\texttt{category name}\}. \\
        & A blurry photo of a \{\texttt{category name}\}. \\
        & A good photo of a \{\texttt{category name}\}. \\
		\hline 
		\multirow{4}{*}{LLM-prompts} &  Describe what a \{\texttt{category name}\} looks like? \\
		& How can you identify a \{\texttt{category name}\}? \\
		& Describe what a \{\texttt{category name}\} in a distance look like within 20 words. \\
		& Describe what a \{\texttt{category name}\} in a low resolution photo look like within 20 words. \\
	\end{tabular}
	\caption{\textbf{The templates and prompts for LLM used in our prompt generator.}}
	\label{tab_cupl}
\end{table*}

\begin{table}[] 
	\centering
	\begin{tabular}{c|ccc}
		\hline 
		Metric & MAE & RMSE \\
		\hline 
		naive template & 25.44 & 42.59 \\
        80 ImageNet templates & 26.54 & 43.82 \\
		general templates & 24.65 & 40.34 \\
        \hline 
        customized prompts & 24.84 & 41.68 \\
        \hline 
        URM-L & 23.83 & 41.03 \\
		\hline
	\end{tabular}
	\caption{\textbf{Ablation of the prompts for language representation.}}
	\label{tab_temp}
\end{table}

\begin{figure}[]
    \centering
    \includegraphics[width=\linewidth]{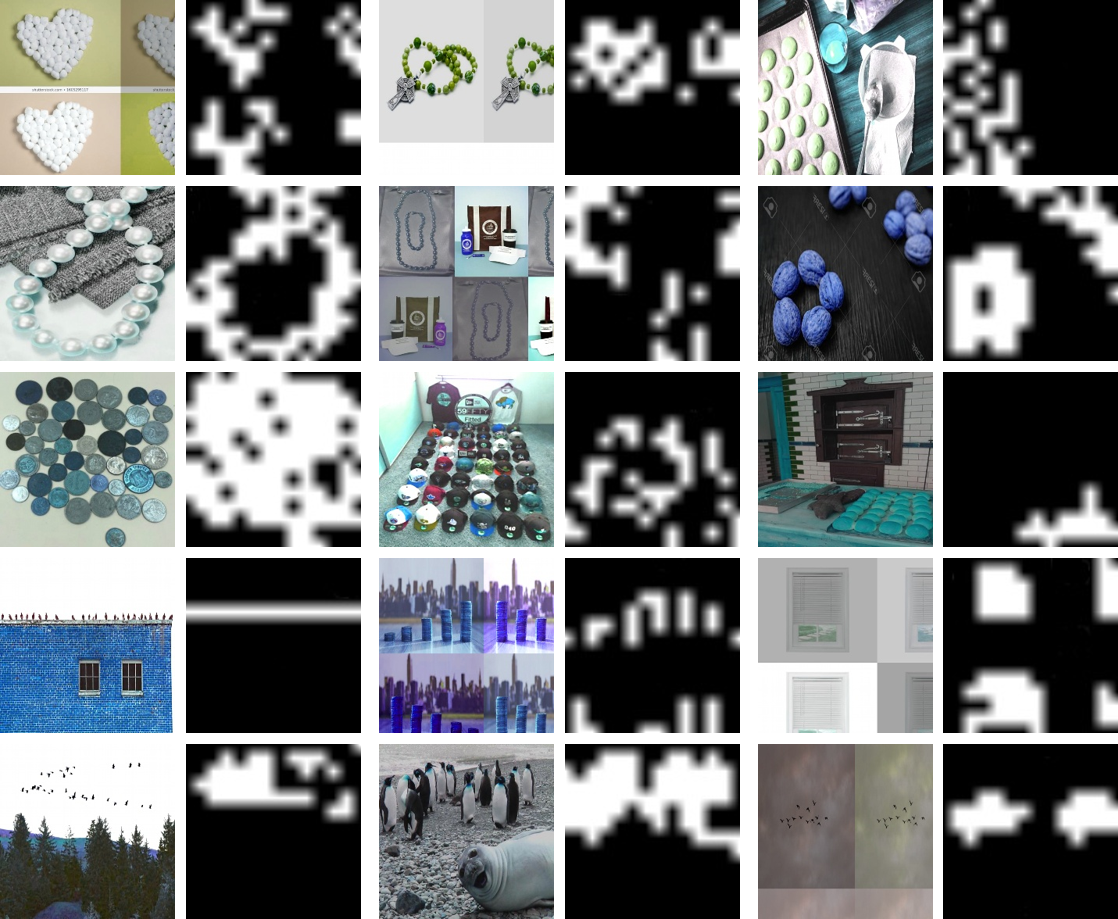}
	\caption{\textbf{Visualization of the segmentation.}}
	\label{fig_seg}
\end{figure}

\section{Visualization}
\label{supp:vis_seg}
The visualization results obtained by MaskCLIP~\cite{Zhou2021ExtractFD} are shown in Figure~\ref{fig_seg}. We show that the method yields compelling open set segmentation results and is robust to data augmentation.

\end{document}